\documentclass[11pt,a4paper]{article}
\usepackage[hyperref]{acl2018_pagenums}
\usepackage{times}
\usepackage{latexsym}
\usepackage{amssymb}
\usepackage{amsmath}
\usepackage[capitalize,nameinlink]{cleveref}
\usepackage{color}
\usepackage{graphicx}
\usepackage{enumitem}
\usepackage{booktabs} 
\usepackage{multirow}
\usepackage{multicol}
\usepackage{subcaption}

\usepackage{url}

\aclfinalcopy 

\crefname{equation}{}{}

\title{Eval all, trust a few, do wrong to none: \\
  Comparing sentence generation models}

\author{Ond\v rej C\' ifka%
  \thanks{\enskip{}Work done during an internship at Google Zurich.} \\
  Charles University \\
  {\tt cifka@matfyz.cz} \\\And
  Aliaksei Severyn \\
  Google Inc. \\
  {\tt severyn@google.com} \\\AND
  Enrique Alfonseca \\
  Google Inc. \\
  {\tt ealfonseca@google.com} \\\And
  Katja Filippova \\
  Google Inc. \\
  {\tt katjaf@google.com}}

\date{}

\begin{document}
\maketitle
\begin{abstract}
In this paper, we study recent neural generative models for text generation related to variational autoencoders.
Previous works have employed various techniques to control the prior distribution of the latent codes in these models, which is important for sampling performance, but little attention has been paid to reconstruction error.
In our study, we follow a rigorous evaluation protocol using a large set of previously used and novel automatic and human evaluation metrics, applied to both generated samples and
reconstructions.
We hope that it will become the new evaluation standard
when comparing neural generative models for text.
\end{abstract}


\section{Introduction}

While automatic natural language generation (NLG), in particular from
structured data, has had a long tradition \cite{reiter00}, the recent
advances in deep learning have given it a new
impetus.
In parallel to a massive number of deep generative models for creating
realistic images, a fair number of papers have introduced
probabilistic generative models of text \cite[inter
alia]{Zhao2017AdversariallyRA,Hu2017TowardCG} which are
claimed to produce fluent and meaningful samples from a
continuous vector representation.
Similar to research focused on image generation, related but distinct text generation
tasks for such models include:
\begin{enumerate}
\item[(1)] sentence reconstruction -- given a natural language sentence,
  can we encode it into a fixed-length vector and then reconstruct it
  from that representation?
\item[(2)] unconditional sentence generation -- can we generate fluent sentences that follow
the distribution of sentences in natural language?
\item[(3)] conditional sentence generation -- given a content and/or style
representation, can we generate a sentence expressing that content and
exhibiting the desired stylistic properties?
\end{enumerate}

While tasks (1) and (2) may not have obvious applications, they are important for assessing
the properties of the learned representations and their usefulness for other tasks, including
(3). For example, autoencoder-based models have been proposed
\cite{Hu2017TowardCG,Zhao2017AdversariallyRA} for learning disentangled representations
of style and content from unaligned data. One possible application of such autoencoders is
modifying the style of a sentence by manipulating the style representation, but this is only
possible if the model can encode and generate accurately, i.e.\ has a low reconstruction error.


Partly due to the difficulty of evaluating text generation directly, recent
studies on autoencoders for text
\cite{Bowman2016GeneratingSF,Hu2017TowardCG} have mostly focused on
applying them to tasks such as language modeling and classification.
With the exception of \citet{Zhao2017AdversariallyRA}, these studies do
not consider the reconstruction task.
\citet{Bowman2016GeneratingSF} report negative results of variational autoencoders
on language modeling, which suggests that the reconstruction error of these
models will be high.

%
The lack of evaluation standards resulted in fierce debates around the
experimental setup of some of the most novel neural-based text generation
studies, to the point that their utility has been
questioned.\footnote{See the discussions around the posts by Yoav
  Goldberg from June 2017: \url{https://medium.com/@yoav.goldberg}.}
This is unfortunate because neural generative models for text do hold real promise for NLG, as the
progress in MT over the past few years has clearly demonstrated.

In this paper, we strive to address the methodological issues with the
current neural text generation research and also close some gaps by
answering a few natural questions to the studies already published.
We focus on neural generative models from the autoencoder family
and their performance on tasks (1) and (2), because we feel that
this is an area that hasn't been sufficiently explored and deserves
a proper treatment before one moves on to more complex setups.

In particular, our contributions are as follows:
\begin{enumerate}
\item We focus on several most recent autoencoder models for sentence
  generation, namely plain (AE), variational (VAE) and adversarially
  regularized (ARAE) autoencoders \cite{Kingma2013AutoEncodingVB,Bowman2016GeneratingSF,Zhao2017AdversariallyRA},
  as well as adversarial autoencoders
  (AAE, \citealp{Makhzani2015AdversarialA}), and compare them on
  equal footing.
\item We study the effects of alternative techniques for
  regularizing autoencoders for text, namely latent code normalization,
  injecting noise into the latent representation, and RNN dropout.
\item We show that these simple techniques are sufficient for training an
  autoencoder which is comparable to state-of-the-art models for
  unconditional text generation while outperforming them in terms of
  reconstruction accuracy.
\item We rigorously evaluate different variants of autoencoder models
  with humans as well as compute a rich set of automatic metrics on
  both generated samples and reconstructions, which is missing in the
  previous work.
\item In particular, we introduce a novel technique for automatically
  measuring the quality of generated texts~--
  \emph{Fr\'echet InferSent Distance} (inspired by
  the recent work on image generation by \citealp{heusel17}).
\end{enumerate}

\section{Related work}
Since the introduction of VAEs by \citet{Kingma2013AutoEncodingVB} and its many successful applications in computer vision, the first study of VAEs for text generation was performed by \citet{Bowman2016GeneratingSF}. While demonstrating that VAEs can be a viable way to train unconditional generative models for text, the authors show that training a VAE with an LSTM~\cite{Hochreiter1997LongSM} decoder leads to an issue where it tends to ignore the latent code completely and hence collapse to a language model. To alleviate this issue, the authors applied two moderately successful tricks: input dropout and KL term annealing. While demonstrating their model can generate natural-looking samples, the reconstruction performance is omitted from the discussion, which is important as it indicates how well the encoder generalizes and structures the latent space.

To address some of the issues of training VAE models for text discussed by~\citet{Bowman2016GeneratingSF}, \citet{Semeniuta2017AHC} propose a hybrid architecture composed of a convolutional encoder and a decoder composed of a de-convolutional and an autoregressive layer (LSTM or ByteNet,~\citealp{ByteNet}). This model is shown to better handle longer sequences and more importantly, it allows for a better control over the KL term. The latter ensures that the latent vector is actually useful and used by the decoder. Additionally, similar to findings in~\citet{Chen16} and~\citet{Yang2017ImprovedVA}, explicit control over the autoregressive power of the decoder, e.g., by using a ByteNet decoder with a smaller receptive field, helps to alleviate this issue. In this work we employ a standard LSTM encoder/decoder architecture, whereas our primary focus is on  various mechanisms to match posterior and prior distributions and its effects on structuring the latent space.

The original VAE objective includes a KL penalty term whose goal is to match the approximate posterior with a prior.  This regularizes (smooths out) the latent space, ensuring that it is possible to generate meaningful samples from any point from the prior. Instead of using a conventional KL penalty, \citet{Makhzani2015AdversarialA} propose to use a GAN discriminator to match the aggregate approximate posterior with the prior. \citet{Vegan} provide a proof that this in effect corresponds to minimizing a Wasserstein distance in the primal between the data and generated distributions. \citet{Zhao2017AdversariallyRA} attack the problem from a different angle by using a GAN to instead learn a powerful prior that matches the aggregated posterior. Thus, during generation, the latent vectors are sampled from the GAN generator instead of being drawn directly from an imposed prior.

\citet{Hu2017TowardCG} propose a conditional VAE model for text where a discriminator is used to impose desired attributes on generated samples and disentangle them from the latent representation produced by the encoder. To enable back-propagation from the discriminator, the recurrent decoder is made fully differentiable by applying a continuous approximation.

Another notable approach of applying VAEs to text was recently proposed in \citet{Guu2017GeneratingSB}, where generation is treated as a prototype-then-edit task -- sample a prototype sentence from the training corpus and then edit it into a new sentence. Unlike conventional VAEs where the encoder packs the whole sentence into a latent vector, \citet{Guu2017GeneratingSB} choose the latent vector to represent an edit that transforms an input prototype into a new sentence.

Finally, autoencoders as a key technique in unsupervised representation learning have been widely applied in NLP tasks to regularize language models and sequence-to-sequence models~\cite{semi-supervised-vae}, for supervised machine translation~\cite{VariationalMT}, and more recently, for enabling unsupervised machine translation~\cite{unsupervised-MT,unsupervised-MT-facebook}.
\section{Background}
In this section, we briefly review two previously proposed
types of generative models which we adopt in this work:
variational and adversarial autoencoders.

Both are autoencoders consisting of two components: an encoder $E$,
which transforms an input $x$ to an embedding (latent code) $z$, and
a decoder (generator) $G$, which produces a reconstruction of $x$ from $z$.
A prior distribution $p(z)$ is imposed on the embedding space and the model
is trained to match the aggregated posterior $p_E(z)=\int_x p_\text{data}(x) p_E(z|x) dx $ to the prior.
The two models differ in the way they achieve this goal: a VAE includes
a KL divergence term in its cost function, while AAEs employ
an adversarial training objective.

\subsection{Variational autoencoder}
A \emph{variational autoencoder} \citep{Kingma2013AutoEncodingVB}
maximizes a lower bound on the marginal log-likelihood:
\begin{equation*}
\begin{split}
\log p_G(x) \geq \mathbb{E}_{p_E(z|x)}[\log p_G(x|z)] \\
- \mathrm{KL}(p_E(z|x) \:||\: p(z)).
\end{split}
\end{equation*}
The first term is the log-probability of reconstructing the input $x$
given the latent vector $z$ sampled from the posterior distribution.
The second term is the negative KL divergence
from the prior to the posterior, which effectively acts as a regularizer, 
pushing the posterior closer to the prior. 

A standard Gaussian is usually chosen as the prior distribution, and the
posterior (the output distribution of the encoder) is modelled as a diagonal
Gaussian to allow for gradient back-propagation using
a reparameterization trick.

A VAE for text, as proposed by \citet{Bowman2016GeneratingSF}, uses an RNN
encoder and decoder. The authors use KL cost annealing
(gradually increasing the weight of the KL term from 0 to 1) and
word dropout (randomly masking out tokens from the decoder's input during
training) to encourage the decoder to make use of the latent vector produced by
the encoder.

\subsection{Adversarial autoencoder}
\emph{Adversarial autoencoders} \citep{Makhzani2015AdversarialA}
regularize the embedding space by means of adversarial training.
The model is extended with an adversarial network (discriminator) $D$,
which is trained to predict whether a given vector $z$ is a sample
from the imposed prior distribution $p(z)$ or an embedding produced by
the encoder:
\begin{equation*}
\begin{split}
\max_{D} \mathbb{E}_{p(z)}[\log p_D(z)] + \\
\mathbb{E}_{p_E(z|x)p_\text{data}(x)}
[\log (1-p_D(z))].
\end{split}
\end{equation*}
Here, $p_D(z)$ is the probability, predicted by the discriminator,
that $z$ is a genuine sample from the prior distribution.

Meanwhile, the autoencoder is trained in two alternating optimization steps:
In the \emph{reconstruction phase}, we optimize the standard reconstruction
objective:
\begin{equation*}
\max_{E,G} \mathbb{E}_{p_E(z|x)p_\text{data}(x)}[\log p_G(x|z)].
\end{equation*}
In the \emph{regularization phase}, the encoder is trained to fool the
discriminator so that the latter is unable to distinguish the encoder
outputs from the samples coming from $p(z)$:
\begin{equation*}
\max_{E} \mathbb{E}_{p_E(z|x)p_\text{data}(x)}[\log p_D(z)].
\end{equation*}

Note that $p(z)$ can now be an arbitrary distribution, as long as we can
sample from it.
\section{Models}
We will now describe the details of the models we examine in this work,
our modifications to them, and our choice of prior distributions.
In general, the models use an LSTM encoder and decoder with 512 units 
and 128-dimensional word embeddings;
the latent code is 100-dimensional.
The discriminator (if present) is a fully connected network with 3 layers of size 300.

\paragraph{VAE.} In the variational autoencoder, we adhere to the
commonly used Gaussian prior and diagonal Gaussian posterior. We employ KL term annealing and word dropout as in \citet{Bowman2016GeneratingSF}.
In one of the settings (\textsc{VAE-bow}), we replace
word dropout with the bag-of-words loss of \citet{Zhao2017LearningDD}.

\paragraph{AAE.} For adversarial autoencoders, we experiment with
two kinds of prior distributions: a standard Gaussian and a uniform
distribution on the unit sphere (in Euclidean space).

In the case of a Gaussian prior, we use two types of posterior
distributions: a diagonal Gaussian parameterized by the encoder, i.e.
$p(z|x) = \mathcal{N}(z;\,\mu_E(x),\,\sigma_E^2(x))$,
and a deterministic posterior, where the encoder produces a single
$z = E(x)$ for each input. We refer to the two resulting models
as \textsc{AAE-gauss} and \textsc{AAE-gauss-det}, respectively.

In the spherical case (\textsc{AAE-sph}),
we normalize the output of the encoder to
ensure that the aggregated posterior distribution is supported on the
unit sphere. During training, we add Gaussian noise
to the normalized embeddings before passing them to the decoder
and the discriminator. The variance of this noise is either fixed
or exponentially decayed over time.

Unlike \citet{Makhzani2015AdversarialA}, we combine the reconstruction and
regularization phase in one training objective:
\begin{equation*}
\max_{E,G} \mathbb{E}_{p_E(z|x)p_\text{data}(x)}
[\log p_G(x|z) + \lambda\log p_D(z)].
\end{equation*}
We use $\lambda=20$ except where stated otherwise.

To further regularize the decoder, we apply dropout with a keep probability
of 0.4 on the LSTM inputs and states~\cite{recurrent-dropout}.

\paragraph{ARAE.} Adversarially regularized autoencoders
\citep{Zhao2017AdversariallyRA} are similar to AAEs,
but instead of imposing a prior distribution on the embeddings,
they learn a flexible prior and employ adversarial training to match it to
the aggregated posterior.
We use the original ARAE implementation,
modified to perform decoding from the mean of the posterior distribution.%
\footnote{Our fork of the code, based on a version from August 2017, can be found here: \url{https://git.io/fxQnz}}
We evaluate two ARAE configurations: the defaults used by
\citet{Zhao2017AdversariallyRA} and a modified setup with hyper-parameters
and training time matching our models.

\paragraph{Plain AE.} We also include a plain autoencoder, which
isn't endowed with a means of controlling the aggregated posterior.
However, in order to be able to draw samples from the model,
we still assume a prior distribution on the embeddings~--
either Gaussian (\textsc{AE-gauss-det})
or spherical (\textsc{AE-sph}).
These autoencoders are equivalent to their adversarial counterparts
with $\lambda$ set to 0.

Note that while there is no explicit control over the embedding space
of \textsc{AE-gauss-det}, the outputs of the \textsc{AE-sph} encoder
are constrained to the unit sphere (although a uniform distribution
is not enforced).
\section{Experimental setup}
We train and evaluate all models on a public corpus consisting of 200,000
sentence summaries extracted from news
articles%
\footnote{\url{https://git.io/fxQnR}} 
\cite{Filippova2013OvercomingTL}.
We perform unsupervised sub-word tokenization using SentencePiece%
\footnote{\url{https://git.io/sentencepiece}}
with a vocabulary of 16,000 tokens.

The models are trained for 500,000 iterations using Adam \citep{Kingma2014AdamAM} with a batch size of 128 and a learning rate of $10^{-3}$.
For AAEs, we use SGD with a learning rate of $10^{-4}$ to update the discriminator.

Each model is evaluated in two different modes:
\begin{itemize}
  \item {\bf Sampling} -- as a pure unconditional generative model, drawing random samples from
    the prior distribution and using them to condition
    greedy decoding.
    This allows us to measure how well the model approximates the
    underlying data distribution.
  \item {\bf Reconstruction} -- as an autoencoder, measuring the
    reconstruction quality. In this case, we first encode the input sentence,
    use the mean of the posterior distribution as the latent vector $z$,
    and then run greedy decoding.
\end{itemize}
We give examples of generated sentences in \cref{sec:examples}.

\subsection{Sampling evaluation}

To evaluate an unconditional generative model for text, we would like to make sure that (a) the
generated sentences are correct with respect to the language used in the
training data, and (b) the generated sentences reflect the diversity of
expressions in the training data, i.e.\ the model avoids mode collapse.
In order to capture both requirements, we use a number of
different evaluation metrics.

\paragraph{Cross entropy.}
A natural way to evaluate a probabilistic model is cross entropy:
\begin{equation}
\mathbb{E}_{p_\text{data}(x)}[-\log p_G(x)].
\label{eq:cross-entropy}
\end{equation}
Note, however, that $p_G(x)=\mathbb{E}_{p(z)}[p(x|z)]$ is
intractable for any
given $x$. Following \citet{Zhao2017AdversariallyRA}, we approximate
$p_G(x)$ using an RNN language model trained on 100,000 model samples;
then, to obtain an estimate of \cref{eq:cross-entropy},
we evaluate this LM on the test set.%
\footnote{Note that \citet{Zhao2017AdversariallyRA} use
an equivalent metric, but refer to it as `reverse perplexity'.}

We are also interested in `reverse cross entropy', i.e.\ the expected
negative log-probability of samples from the model with respect to the
true data distribution:
\begin{equation}
\mathbb{E}_{p_G(x)}[-\log p_\text{data}(x)].
\label{eq:rev-cross-entropy}
\end{equation}
This can be thought of as a measure of plausibility (fluency) of the generative model's outputs. Again, $p_\text{data}(x)$ is unknown, but can
be approximated using a language model.
Therefore, to estimate \cref{eq:rev-cross-entropy},
we score the samples from each model using a pre-trained RNN LM.
The model is trained on a large news corpus from the English Gigaword.\footnote{\url{https://catalog.ldc.upenn.edu/ldc2003t05}}

\paragraph{Fr\'echet distance.}
In addition, motivated by a comprehensive study of various GAN models~\citep{gans-study} where the authors use Fr\'echet Inception Distance \citep{heusel17} extensively and demonstrate that it is superior to the Inception Score~\cite{incepction-score}, we experiment with an equivalent metric for text~-- \emph{Fr\'echet InferSent Distance} (FID)~-- to measure the distance between the generative distribution and the data distribution.
FID measures the Wasserstein-2 distance \cite{Vaserstein1969Markov} between two Gaussians,
whose means and covariances are taken from embeddings of the real and generated data (i.e.\ samples from $p_\text{data}$ and $p_G$), respectively.
To our knowledge, this is the first time that this idea is applied to evaluating generative models for text. Different from the negative log-likelihood metrics discussed above, FID directly measures the distance between distributions, hence it offers an additional angle for comparing generative models whose goal is to learn to recover the true data distribution.

We compute FID between 10,000 sentences generated from the model and taken from the test set, respectively. To obtain their embeddings, we use a pre-trained general purpose sentence embedding model, InferSent~\cite{infersent}, which encodes each sentence as a 4,096-dimensional vector. 
We chose InferSent for computing the FID metric on sentence samples because it has been shown to provide state-of-the-art results on various sentence representation tasks, and is domain-independent to a large extent.

\subsection{Reconstruction evaluation}
The metrics in the previous section quantify the quality and diversity of samples generated while conditioning the decoder on a sample from the prior $p(z)$. Another way to gauge the diversity of sentences the model can represent is to measure how accurately it can reconstruct a given input.
We express the reconstruction error as negative log-likelihood
(NLL) and BLEU-3 and ROUGE-3 scores computed with the
input sentence as a reference.

\subsection{Human evaluation}
We also evaluate the models based on subjective human judgment, focusing on the two tasks mentioned above: sampling and reconstruction.

For sampling, we decoded sentences from random points in the embedding space
and asked human raters to rate them on a 5-point Likert scale according to
their fluency, where 1~= gibberish, 3~= understandable but
ungrammatical sentences, and 5~= naturally constructed and
understandable sentences.

For reconstruction, we presented the raters with a sentence and its reconstruction produced by one of the models. Besides assessing fluency, the raters were asked to provide another score on a 5-point Likert scale measuring how well the output reflects the
original meaning (this score is referred to as relevance in the following). A score of 1 corresponds to an unrelated sentence,
3~to a reasonably good paraphrase, and 5 means either a perfect
reconstruction or a semantically equivalent paraphrase.

The evaluation was done using a crowdsourced rating platform.
For both tasks, we evaluated a sample of 200 sentences from each model,
employing three raters per item.
The results were calculated as an average of the median sentence ratings.
For 84\% of the items, there was a majority score, i.e.\ at least two of the three raters chose the same of the 5 possible scores for the item.

\section{Results}
\begin{table*}
\begin{center}
\makebox[\textwidth]{
\begin{tabular}{lrrrrrr}
\toprule
\multirow{2}{*}{\bf Model} & \multicolumn{3}{c}{\bf Sampling}  & \multicolumn{3}{c}{\bf Reconstruction} \\
\cmidrule(lr){2-4} \cmidrule(l){5-7}
 & \multicolumn{1}{c}{\bf Forward} & \multicolumn{1}{c}{\bf Reverse} & \multicolumn{1}{c}{\bf FID}  & \multicolumn{1}{c}{\bf BLEU}  & \multicolumn{1}{c}{\bf ROUGE}  & \multicolumn{1}{c}{\bf NLL} \\
\midrule
real data & 73.11 & 75.38 & 0.4193  & \multicolumn{1}{c}{---} & \multicolumn{1}{c}{---} & \multicolumn{1}{c}{---} \\

\textsc{LM} & 78.50 & 88.75 & 0.6267  & \multicolumn{1}{c}{---} & \multicolumn{1}{c}{---} & \multicolumn{1}{c}{---} \\

\textsc{VAE} ($d_{\mathrm w}=0.5$) & 79.75 & 65.46 & {\bf 0.6562} & 10.27 & 20.52 & 66.3 \\

\textsc{AAE-sph} ($\sigma=0.075$) & {\bf 74.01} & 82.34 & 0.6622 & 50.90 & 59.93 & 36.0 \\

\textsc{AAE-sph} ($\sigma=0.1$) & 76.28 & 66.73 & 0.6632 & 35.19 & 46.53 & 42.1 \\

\textsc{AE-sph} ($\sigma=0.1$) & 75.77 & 67.98 & 0.6635 & 37.56 & 49.42 & 26.3 \\

\textsc{AAE-sph} ($\sigma=0.05$, $\lambda=10$) & 74.28 & 103.33 & 0.6749 & 60.39 & 68.14 & 31.3 \\

\textsc{AE-sph} ($\sigma=0.05$) & 74.69 & 101.27 & 0.7403 & 63.22 & 70.08 & 14.3 \\

\textsc{ARAE} & 80.48 & 94.51 & 0.7871  & {\bf 72.21} & 75.11 & {\bf 7.1} \\

\textsc{AE-sph} ($\sigma\to0$, $d=1$) & 79.62 & 117.911 & 0.8748 & 11.75 & 20.80 & 105.8 \\

\textsc{ARAE} (default) & 99.67 & 73.37 & 0.8860  & 18.38 & 26.08 & 33.4 \\

\textsc{VAE-bow} ($d_{\mathrm w}=1$) & 87.75 & 63.06 & 1.0150 & 2.03 & 11.43 & 88.9 \\

\textsc{AAE-gauss-det} ($\lambda=10$, $d=1$) & 88.49 & 116.18 & 1.1433 & 68.01 & 73.29 & 23.6 \\

\textsc{VAE} ($d_{\mathrm w}=0.75$) & 112.16 & {\bf 59.59} & 1.2440 & 2.04 & 11.30 & 70.5 \\

\textsc{AE-gauss-det} ($d=1$) & 107.16 & 71.12 & 3.0839 & 71.14 & {\bf 76.25} & 9.9 \\


\bottomrule
\end{tabular}
}
\end{center}
\caption{Automatic evaluation results.
\textbf{Forward}: `forward cross entropy', i.e.\ the negative log-likelihood (NLL) of a LM trained on the
samples from each model and evaluated on the test set;
\textbf{Reverse}: `reverse cross entropy', i.e.\ the NLL of a LM trained on real data and
evaluated on the model samples;
\textbf{FID}: Fr\'echet InferSent Distance.
The reconstruction section reports the
\textbf{NLL}, \textbf{BLEU} and \textbf{ROUGE} w.r.t.\ the
input sentence.
$\sigma$ denotes the standard deviation of the noise added to the sentence embeddings during training. $d$ and $d_{\mathrm w}$ denote the RNN dropout and word dropout keep probability in the decoder, respectively ($d=1$
means no dropout).
`Real data' corresponds to samples from the training set.
}
\label{table:nll_on_eval_set}
\end{table*}

\begin{table*}
\begin{center}
\begin{tabular}{lrr}
\hline
{\bf Model} & {\bf Relevance} & {\bf Fluency} \\
\hline
real data                           & --- & 4.42 \\
\textsc{AAE-gauss-det} ($\lambda=10$, $d=1$) & {\bf 3.54} & 3.71 \\
\textsc{ARAE}                       & 3.35  & 3.56 \\
\textsc{AAE-sph} ($\sigma=0.075$)   & 2.76 & 3.53 \\
\textsc{AE-sph} ($\sigma=0.1$)      & 2.54 & 3.53 \\

\textsc{AAE-sph} ($\sigma=0.1$)     & 2.40 & 3.54 \\
\textsc{AE-sph} ($\sigma\to0$, $d=1$)&1.73 & 2.33 \\
\textsc{ARAE} (default)             & 1.48 & 2.51 \\
\textsc{VAE}  ($d_{\mathrm w}=0.5$)            & 1.39 & {\bf 3.87} \\
\hline
\end{tabular}
\end{center}
\caption{Human evaluation results for the reconstruction task.
Each score is on a scale of 1 to 5. The readability score for
real data from \cref{table:human_sampling} is included for comparison.}
\label{table:human_reconstruction}
\end{table*}

\begin{table}
\begin{center}
\begin{tabular}{lr}
\hline
{\bf Model} & {\bf Fluency} \\
\hline
real data                           & 4.42 \\
\textsc{VAE} ($d_{\mathrm w}=0.5$)  & {\bf 3.46} \\
\textsc{AE-sph} ($\sigma=0.1$)      & 3.07 \\
\textsc{AAE-sph} ($\sigma=0.1$)     & 2.83 \\
\textsc{LM} 						& 2.69 \\
\textsc{AAE-sph} ($\sigma=0.075$)   & 2.61 \\
\textsc{ARAE} (default)             & 2.08 \\
\textsc{AE-sph} ($\sigma\to0$, $d=1$)&1.85 \\
\textsc{ARAE}                       & 1.68 \\
\textsc{AAE-gauss-det} ($\lambda=10$, $d=1$) & 1.53 \\
\hline
\end{tabular}
\end{center}
\caption{Human evaluation results for the sampling task. Each score is on a
  scale of 1 to 5.}
\label{table:human_sampling}
\end{table}

\subsection{Quantitative evaluation}
The results are shown in \cref{table:nll_on_eval_set}
(automatic evaluation) and \cref{table:human_sampling,table:human_reconstruction}  (human evaluation).
Results on samples from the training
set and from an RNN LM are included
for comparison. The RNN LM uses the same architecture and hyperparameters
as the decoder of all the other models.

For the sampling task, one thing to notice is that there seems to be a
trade-off between the quality and the diversity of the samples: models with
a lower (i.e.\ better) reverse cross entropy and a higher fluency
rating tend to have a higher (i.e.\ worse) forward cross entropy.
In particular, the reverse cross entropy of some models
(\textsc{VAE} and some \textsc{-sph} models)
is less than that of the real data~-- this is a clear sign that the model is
suffering from a mode collapse. This is supported by the fact that these
models also tend to have worse performance on reconstruction, which
suggests that the set of sentences they are able to encode is less diverse.

Another important observation is that plain autoencoders with the
spherical prior (\textsc{AE-sph}) achieve relatively good results,
on par with their adversarial counterparts (\textsc{AAE-sph}).
This suggests that the techniques applied in these models~-- constraining
the embeddings to lie on a unit sphere and injecting noise~-- are
sufficient for making the model learn to cover the sphere uniformly and
be able to decode sentences from any given point on
the sphere. The adversarial training seems to have little additional
effect, if any at all.

In particular, \textsc{AE-sph} with $\sigma=0.1$ 
performs at least as well on sampling as all other types of
models we evaluated:
\begin{itemize}
	\item It achieves a superior forward cross entropy.
	\item Its FID is only slightly higher than for
		VAE ($d_{\mathrm w}=0.5$), which achieves the lowest
		(i.e.\ best) value.
	\item Although its reverse cross entropy is still below the real
		data threshold, it is higher than for VAEs, hence it
		arguably suffers less from the mode collapse problem.
	\item It achieves a higher fluency score than a LM and is only
		surpassed by the VAE.
	\item Finally, it outperforms VAEs on the reconstruction task by
		a large margin.
\end{itemize}

The effect of adversarial training on the Gaussian
prior model (\textsc{AAE-gauss-det}) seems to be more pronounced
than in the spherical prior models~--
this is unsurprising as the non-adversarial variant (\textsc{AE-gauss-det})
doesn't place any restrictions on the aggregated posterior, and therefore
cannot be expected to be useful as a generative model. However,
\textsc{AAE-gauss-det} still has poor performance on
sampling according to both automatic and human evaluation.

Regarding \textsc{ARAE}, it outperforms all other methods on almost all
reconstruction metrics, but its results on sampling are rather poor, especially
according to human ratings. This might be due to a more challenging dataset
than in \citet{Zhao2017AdversariallyRA}, or simply because of the model's
high sensitivity to hyperparameters, which is noted by the authors.

\subsection{Embedding visualization}
\begin{figure*}
\begin{subfigure}[b]{0.5\textwidth}
\centering
\includegraphics[width=5.7cm,trim={2.4cm 2cm 2cm 1.3cm},clip]{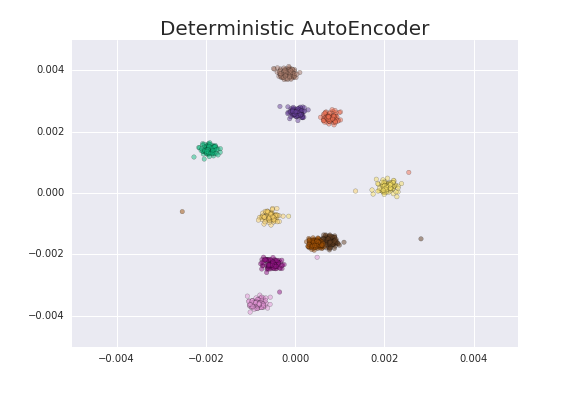}
\caption{\textsc{AAE-gauss-det}}
\end{subfigure}
\begin{subfigure}[b]{0.5\textwidth}
\centering
\includegraphics[width=5.7cm,trim={2.4cm 2cm 2cm 1.3cm},clip]{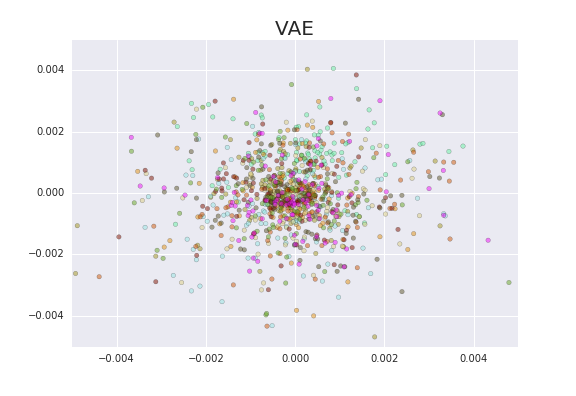}
\caption{\textsc{VAE}}
\end{subfigure}
\begin{subfigure}[b]{0.5\textwidth}
\centering
\includegraphics[width=5.7cm,trim={2.4cm 2cm 2cm 1.3cm},clip]{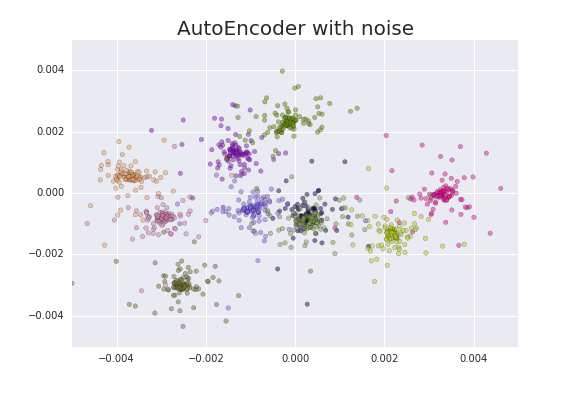}
\caption{\textsc{AE-sph}}
\end{subfigure}
\begin{subfigure}[b]{0.5\textwidth}
\centering
\includegraphics[width=5.7cm,trim={2.4cm 2cm 2cm 1.3cm},clip]{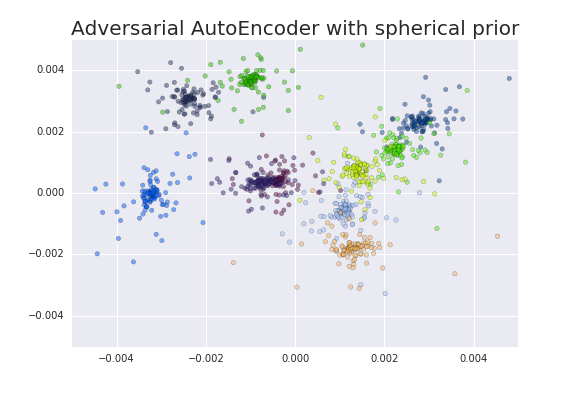}
\caption{\textsc{AAE-sph}}
\end{subfigure}
\caption{t-SNE visualization of 100 different encodings (samples from
  the posterior distribution) of 10 sentences, using various models.}
\label{fig:encodings}
\end{figure*}

\cref{fig:encodings} shows t-SNE~\cite{tsne} projections in 2D of the encodings of ten
random sentences from the test set. Each sentence has been encoded one
hundred times with sampling from the posterior, then plotted
with some additional noise in order to
better visualize collapsed points.

Plain {\sc AE-gauss-det} is deterministic, and each sentence is
mapped to the same identical point all 100 times. This leads to a
very high-quality reconstruction, but the embedding space is not
smooth and sampling from random points in the prior would often produce unreadable
outputs. Plain VAE exhibits the opposite behaviour: all 10 inputs are
encoded into large, heavily overlapping regions of the embedding space. This
hints at why this model performs poorly for reconstruction and has
very good quality sampling from any random point in the embedding
space.

Finally \textsc{AAE-sph}
and \textsc{AE-sph} display similar behaviours, with sentences mapped
into smooth regions in the space without significant overlap in the
projections.

While not a quantitative study by itself, the plots are consistent
with the observed results for sampling and reconstruction described above.

\section{Conclusions}

We introduced a rigorous evaluation scheme for generative models for text.
In addition to previously proposed metrics, we proposed the
Fr\'echet InferSent Distance, adopted from the field of image generation.

Three families of generative models (plain, variational and
adversarially regularized autoencoders) have been thoroughly compared,
under different regularization strategies. The qualitative evaluation
shows that no model outperforms the others under all circumstances,
with \textsc{VAE} being the strongest for sampling, but suffering from
mode collapse and poor reconstruction performance. The rest of the
models represent compromises between good sampling and reconstruction,
and as we have demonstrated, the trade-off between these two can be
controlled using simple regularization techniques.

\bibliographystyle{acl_natbib}
\bibliography{bibliography}

\newpage
\appendix
\onecolumn
\section{Example outputs}
\label{sec:examples}
\cref{fig:samples} shows a random sample of the sentences generated from various models.

In \cref{fig:interpolations1,fig:interpolations2}, we show the outputs obtained by encoding two sentences from the test set and using spherical linear interpolation (Slerp; \citealp{Shoemake1985AnimatingRW}) to generate sentences `in between'.

\begin{figure*}[ht]
\begin{multicols}{2}
\small
\textbf{\textsc{VAE} ($d_{\mathrm w}=0.5$)}\\
Stocks headed for a lower straight morning loss.\\
Justin Bieber is to be disclosed ambitious.\\
Barack Obama has called for a semeter test.\\
A chiropractor has been charged with altering a drug overdose.\\
Bob Grimes has been named a coach of the year.\\
An Arab citizen killed his wife was sentenced to prison.\\
Tommy Wiliser has not able to re-launchs called out.\\
The LVMET has launched a new album, and tour dates.\\
Whose body search for the longest poll wiring at a faulty hour.\\
General Motors has announced a recall of its Sonata sedans.\\
Jaswant Singh has been suspended indefinitely.\\
I 'm.\\
Police are looking for a man who was shot in the police vehicle.\\
The earthquake of Argos.\\
A police officer shot and killed a man.\\
Nicole Richie is a lot of his swipe out out of his childhood collection.\\
Asman Paul McCartney is excited to play in the City Hall.\\
Aust stocks rose, tracking Yankee lifted.\\

\textbf{\textsc{AAE-sph} ($\sigma=0.1$)}\\
Art Group Inc.'s planning to cut its performance on performances.\\
Harbentin's going to make the Broadway.\\
Josey Gucci will be coming to the Liberal president.\\
A New York woman was charged with threatening to hit a hospital on chest in Christie.\\
Idea 4 has finally been launched in Naiabad.\\
A man was charged with stabbing his wife.\\
Barack Obama has found peguch.\\
India is gaining its range Agnimel, a maritime maritime serial cancerist.\\
Jautag mobile phone service hours will be associated.\\
Zac Efron was, but he has lost the Showtime Ryder Cup.\\
But Israel are having a serious mistake.\\
The Diamond Bay Packers have agreed to pitch in the Midwest League.\\
France is the UN Security Council for a UNESCO service.\\
Staff of Staff has been put on their greatest brand.\\
Chift is facing asylum.\\
FoodCorp, will spend for more quarterly.\\
The Tucson inquiry has recovered a new contract.\\
Processs are set to play for gay marriage.\\
REXI is expanding its first profit to the NASDAQ package.\\
Mobile businesses say a profit rise.\\

\textbf{\textsc{AE-sph} ($\sigma=0.1$)}\\
A man was killed and killed in a motorcycle crash.\\
Two remomos die 230,000 tons of wheat from an bayevogue.\\
Freebes Wainwright will be surgery to visit Cronulla.\\
Carl Pro-Arts will appear on hold talks with Golden Boy.\\
The New York Yankees will celebrate the company's new crystal-life company's new Generation Officer.\\
PFDR has reported its biggest multi-tax profit.\\
A quarter ended today was completed from Portum.\\
Lincoln funeral leaders, said he's retiring.\\
Harry Styles has been addicted as a scientist.\\
Likov is aiming to take off the industry.\\
Orlando Magic $\langle$UNK$\rangle$ 4 has won the Red Sox team.\\
A French cruise ship has erupted over a South Ham.\\
DYSE was arrested for stealing $\langle$UNK$\rangle$ 192,000, scoitizing.\\
The team of youngsters are being urged to join the country's best lifestyle.\\
Easy Arctic, is looking to hit the back of the iPhone 4.\\
First Foods Inc. has acquired a strike for non-core offices in Athens.\\
World Home has demanded the holiday season.\\
But does not be screened to the screen of the iPad.\\
This insurance would the health care agency for a year.\\
The Bobcats has hit the ban on the start of the San Jose Earthquakes.\\

\textbf{\textsc{ARAE}}\\
Marvin Armstrong has been impressed.\\
The running in Angelina walked control up, the life.\\
irburg spoke to over a 6 An state is showing Kanover mayor.\\
A Wayne James man, accusing The Pittsburgh sent no fund at his time to reang 2014.\\
Craig said being is planning the United States, admitted she are, placed CBo Sangan.\\
Hosting Goldman Micherro set team at school home.\\
The A woman 24, is gun with.\\
E have lawsuit by a successful court declared to Fire.\\
The Forest Service won to Bruce Wrightan Tur in Ciles.\\
A mobile pig is facing revamp Project Nat out He..\\
Bill Don, Mc Milan ' calls for the athletic history.\\
A Ensaling create a port in the cell and burning north of single County.\\
The Navy is winning to start the Globe.\\
The skulls program sold its vehicle message to New Ten railways.\\
Chinese Illinois burned emergency inpower.\\
Papersa Ltd. plans to sell its open local homes, luxury.\\
Pacific Industries lost manufacturing capacity.\\
80 tube flags set toll run.\\

\end{multicols}
\caption{Random samples from different models.}
\label{fig:samples}
\end{figure*}

\begin{figure*}[ht]
\begin{center}
\textbf{\textsc{VAE} ($d_{\mathrm w}=0.5$)}\\
A man who was shot while walking through a tree. \\
A man has died after falling from a bridge. \\
A man has died after falling into ice. \\
A man was shot dead in his home. \\
A man was shot during a fight. \\
A man was stuck on the face. \\
Barcelona have made a winning start. \\
Barcelona have made a winning start. \\
Barcelona will face a winning start. \\
Barcelona will face a winning start. \\[2ex]

\textbf{\textsc{AAE-sph} ($\sigma=0.1$)}\\
A Doncaster man suffered injury after a collision with a train. \\
A Doncaster man suffered injury after a crash was shot. \\
A Doncaster man is facing charges after a police officer. \\
A Doncaster man faces charges after a domestic assault. \\
A Doncaster man faces charges after he was shot. \\
Barcelona will appeal the suspension of heavy losses. \\
Barcelona will appeal the suspension of suspension. \\
Barcelona will appeal the suspension of suspension. \\
Barcelona will appeal the transfer ban. \\
Barcelona will appeal the transfer ban. \\[2ex]

\textbf{\textsc{AE-sph} ($\sigma=0.1$)}\\
A man suffered life after suffering a tree. \\
A man suffered life after suffering a car. \\
A man suffered life after suffering a car. \\
A man suffered life after suffering a car. \\
A man suffered injury after a car crash. \\
A man will not leave the road. \\
Barcelona will not appeal the transfer. \\
Barcelona will appeal the transfer ban. \\
Barcelona will appeal the transfer ban. \\
Barcelona will appeal the transfer ban. \\
\end{center}
\begin{tabular}{p{0.5\textwidth}p{0.5\textwidth}}

\end{tabular}
\caption{Sentences generated by interpolating between the encodings of ``A Doncaster man suffered life threatening injuries after a collision.'' and ``Barcelona will appeal the transfer ban.''.}
\label{fig:interpolations1}
\end{figure*}

\begin{figure*}[ht]
\begin{center}
\textbf{\textsc{VAE} ($d_{\mathrm w}=0.5$)}\\
A woman is accused of stealing a trio of fireworks. \\
A woman was arrested for stealing a laptop to charity. \\
A woman was found guilty of stealing from her home. \\
A woman has been charged with a string of burglaries. \\
A woman has been charged with a string of burglaries. \\
A school principal was arrested for a string of burglaries. \\
A school principal has been convicted of a federal tax scheme. \\
The Australian Open has been voted as a new Limerick person. \\
The Australian Open has been launched in a new reality show. \\
The Humane Society has been awarded a new distribution centre. \\[2ex]

\textbf{\textsc{AAE-sph} ($\sigma=0.1$)}\\
A woman is accused of leaving her children at home.\\
A woman is accused of leaving her children at home.\\
A woman is accused of leaving her children at a home.\\
A woman is accused of leaving a children at her home.\\
A woman has been accused of leaving a children's home.\\
A woman has been charged with a student at a school bus.\\
The Union has been named a new national national school.\\
The Union has been named a national national national national site.\\
The Union has been named a national national national national site.\\
The Union has been named a national national national national site.\\[2ex]

\textbf{\textsc{AE-sph} ($\sigma=0.1$)}\\
A woman is accused of leaving her home in home.\\
A woman is accused of leaving her home in her home.\\
A woman is accused of leaving her home in her home.\\
A woman is accused of leaving her children in her home.\\
A woman has been accused of vandalizing her children.\\
A woman has been accused of vandalizing her child.\\
The Union has declared a heritage partnership with the city.\\
The Union has been declared a heritage site.\\
The Union has been declared a heritage site.\\
The Union has been declared a heritage site.\\
\end{center}
\begin{tabular}{p{0.5\textwidth}p{0.5\textwidth}}

\end{tabular}
\caption{Sentences generated by interpolating between the encodings of ``A woman is accused of leaving her children at home.'' and ``The Union Buildings have been declared a national heritage site.''.}
\label{fig:interpolations2}
\end{figure*}

\end{document}